\documentclass{article}
\usepackage{ijcai22}

\usepackage{times}
\usepackage{soul}
\usepackage{url}
\usepackage[hidelinks]{hyperref}
\usepackage[utf8]{inputenc}
\usepackage[small]{caption}
\usepackage{graphicx}
\usepackage{amsmath}
\usepackage{amsthm}
\usepackage{booktabs}
\usepackage{algorithm}
\usepackage{algorithmic}
\usepackage{natbib}  
\usepackage{amsmath}
\usepackage{amsfonts}
\usepackage{amssymb}
\usepackage{tikz}
\usepackage{tikz}

\usepackage{latexsym}

\usepackage{algorithm}
\usepackage{algorithmic}
\usepackage{booktabs}  
\usepackage{graphicx} 
\usepackage{subcaption} 
\usepackage{color}
\newcommand{\sysname}{NIAGRA} 

\usepackage{newfloat}
\usepackage{listings}
\DeclareCaptionStyle{ruled}{labelfont=normalfont,labelsep=colon,strut=off} 
\floatstyle{ruled}
\newfloat{listing}{tb}{lst}{}
\floatname{listing}{Listing}
\usepackage{tikz}
\def\checkmark{\tikz\fill[scale=0.4](0,.35) -- (.25,0) -- (1,.7) -- (.25,.15) -- cycle;} 
\title{An Ensemble Approach for Automated Theorem Proving Based on \\ Efficient Name Invariant Graph Neural Representations}

\author{
    Achille Fokoue, Ibrahim Abdelaziz, Maxwell Crouse, Shajith Ikbal, \\ 
    {\bf Akihiro Kishimoto, Guilherme Lima, Ndivhuwo Makondo, Radu Marinescu }
    \emails
    achille@us.ibm.com, 
    \{ibrahim.abdelaziz1, maxwell.crouse, akihiro.kishimoto, guilherme.lima, ndivhuwo.akondo\}@ibm.com, 
    shajmoha@in.ibm.com, 
    radu.marinescu@ie.ibm.com
    \affiliations
    IBM Research
}

\begin{document}

\maketitle

\begin{abstract}
Using reinforcement learning for automated theorem proving has recently received much attention. Current approaches use representations of logical statements that often rely on the names used in these statements and, as a result, the models are generally not transferable from one domain to another. The size of these representations and whether to include the whole theory or part of it are other important decisions that affect the performance of these approaches as well as their runtime efficiency. In this paper, we present {\sysname}; an ensemble \underline{N}ame \underline{I}nv\underline{A}riant \underline{G}raph \underline{R}epresent\underline{A}tion. {\sysname} addresses this problem by using 1) improved Graph Neural Networks for learning name-invariant formula representations that is tailored for their unique characteristics and 2) an efficient ensemble approach for automated theorem proving. Our experimental evaluation shows state-of-the-art performance on multiple datasets from different domains with improvements up to 10\% compared to the best learning-based approaches. Furthermore, transfer learning experiments show that our approach significantly outperforms other learning-based approaches by up to 28\%.
\end{abstract}

\section{Introduction}


At its core, automated theorem proving is a complex search problem, wherein the construction of a proof is treated as a search through sound inference steps until a satisfactory conclusion is derived. While proof search is a generally intractable problem, significant amounts of research devoted towards the design and implementation of domain-specific search heuristics have allowed automated theorem provers (ATPs) to function effectively in application areas spanning from distributed systems \cite{hawblitzel2015ironfleet, garland2000using} to medicine \cite{masci2014formal,lucas1993representation}. However, as the set of domains that ATPs have been applied to has grown, the need to move away from hand-crafted search heuristics has been recognized \cite{ARCADE2017:We_know_nearly_nothingl}.


Recently, there has been strong interest in applying deep learning to proof search 
\cite{paliwal2020graph,bansal2019icml,loos2017lpar, crouse2021deep}, with the ultimate goal being methods capable of guiding proof search that can automatically adjust themselves to a particular domain with little-to-no manual feature engineering. The earliest deep learning-based approaches \cite{loos2017lpar} to proof guidance operated over shallow representations of their inputs, where logical formulas were treated as bags of symbols. While able to achieve impressive results at the time, these methods fell short against later approaches designed to capture the rich graph structure inherent to logical formulas. 


With the advent of geometric deep learning \cite{bronstein2017geometric}, methods that could explicitly account for the structural characteristics of logical formulas quickly began to gain traction. Early usages of graph neural networks (GNNs) for theorem proving often focused on offline tasks like premise selection \cite{wang2017neurips, crouse2019arxiv, rawson2020directed}, as the added computational expense of graph neural networks outweighed any value they could provide when used during proof search. It has only been quite recently that works incorporating graph neural networks directly into purely neural proof guidance \cite{paliwal2020graph, abdelaziz2022learning, aygun2022proving} have been able to demonstrate state-of-the-art results. 



\begin{figure}[t!]
\begin{center}
\includegraphics[width=0.7\columnwidth]{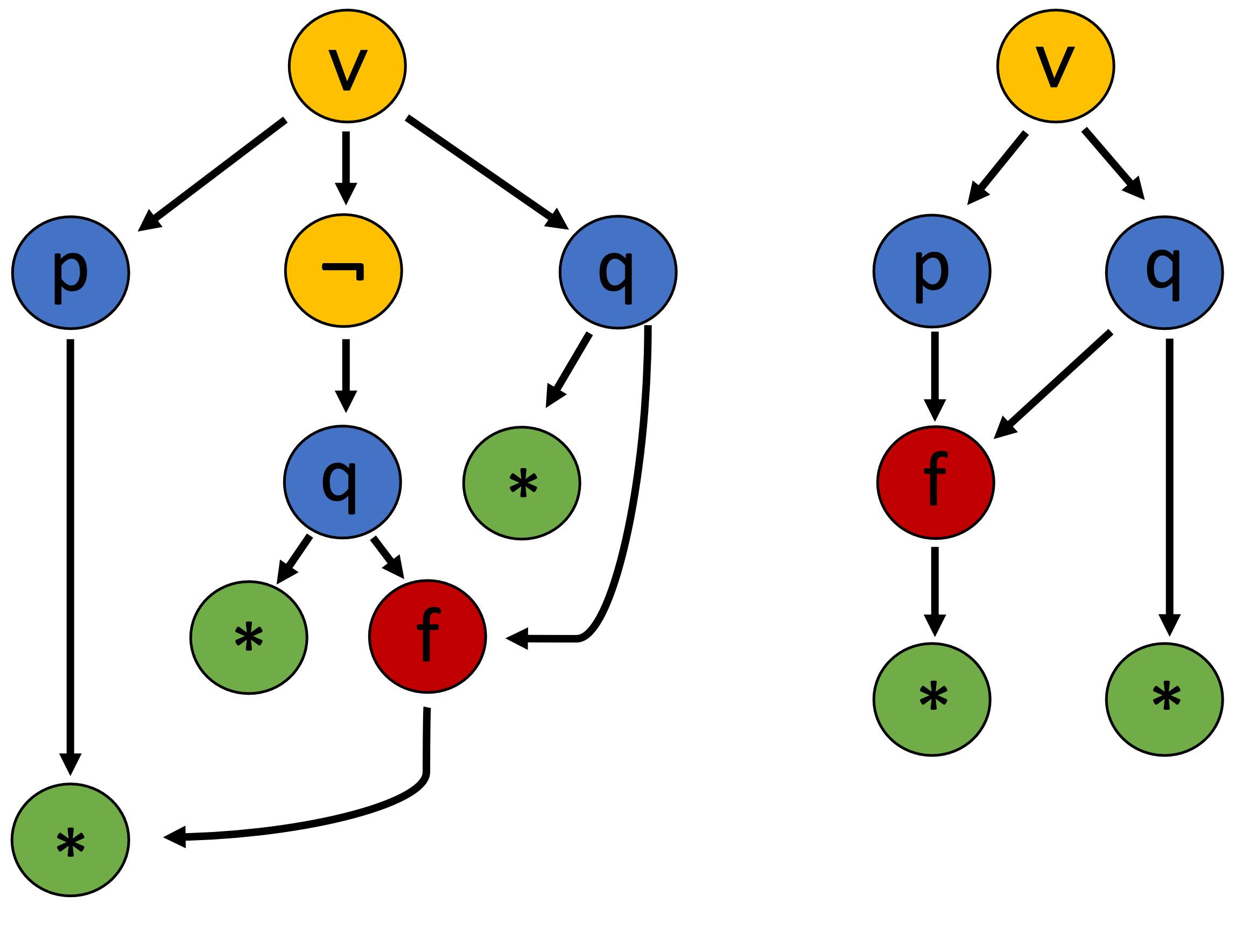}
\caption{DAG representations for two clauses, $\forall A, B, C .\ p(A) \vee \lnot q(B, f(A)) \vee q(C, f(A))$ and $\forall X, Y.\ q(f(X), Y) \vee  p(f(X))$, with anonymized variables to ensure renaming invariance.}
\label{fig:dag_example}
\end{center}
\end{figure}

Most state-of-the-art deep learning-based proof guidance systems leverage graph representations of logical formulas in roughly the same way. These systems first convert logical formulas into directed acyclic graphs (DAGs). They then apply any of the many GNN variants to each formula's DAG representation to produce a dense, real-valued vector that can be processed by other neural components. These approaches convert each logical formula into a separate graph that is disconnected from the graphs representing other formulas in the same theory (an example of this is shown in Figure~\ref{fig:dag_example}). When these isolated graphs are processed by GNNs, they are thus processed independently of one another. This simple approach is very efficient in terms of run-time performance and computational overhead for two main reasons: (1) it enables the parallel computation of each formula's graph embedding, and (2) it enables the reuse of those formula embeddings within a proof attempt, since a formula's graph will remain unchanged throughout the reasoning process. Though attractive from an efficiency perspective, in general, this approach is not faithful to the semantics of logical formulas. This is because the meaning of a formula in a particular proof attempt is a function of all other formulas available to the reasoner. 



In order to preserve the inter-dependencies between formulas, it would be more appropriate to represent the set of all formulas with a single, connected DAG. In practice, this is too expensive (in terms of both memory and computation time) to execute within modern, highly optimized theorem provers. Prior work \cite{jakubuuv2020enigma} exploring more holistic graph embeddings found a middle ground by batch processing subsets of formulas together. That is, their approach would aggregate unprocessed clauses until a size threshold was reached and then embed the aggregated set of clauses together with a GNN. While effective, their approach (1) introduces a delay between when a clause was inferred and when it was embedded, (2) requires frequent large-scale graph embedding operations, and (3) computes only a single, fixed relevance score for each clause.

In this work, we introduce {\sysname} (\underline{N}ame \underline{I}nv\underline{A}riant \underline{G}raph \underline{R}epresent\underline{A}tion). {\sysname} aims to achieve a representational fidelity that surpasses previous works while simultaneously avoiding the significant computational expense involved in repeatedly computing graph embeddings of the evolving large connected graph representing the state of the prover (i.e., the graph containing the representations of the original axioms and the derived formulas). {\sysname} achieves this through a novel embedding process that, prior to proof search, uses the initially provided theory to generate embeddings for individual non-logical symbols (i.e., constant, function and predicate names), that are representative of the complete structural context in which the symbols occur. Then, during theorem proving the embeddings for formulas are computed independently (as with prior approaches), but using the theory-level symbol embeddings as the graph's initial node embeddings.

In addition to our main contribution of a novel graph embedding process, we also consider and evaluate alternative ensembling schemes for neural-guided theorem proving. In particular, we show how standard ensembling methods can be improved by taking into account not only the weights of the neural architecture, but also the configuration of the underlying theorem prover that executes inference steps.

Through extensive experimentation, we find that this approach yields state-of-the-art performance as compared to other unsupervised deep learning-based and standard theorem provers on standard benchmark datasets. Notably, on the hard subset of  Mizar~\cite{grabowski2010mizar} (MPTP), it significantly outperforms the best unsupervised learning-based system by 10\% (HER:~\cite{aygun2022proving}) and the best classical ATPs by 3\% (Vampire) and 18\% (E), and transfer learning experiments show that our approach significantly outperforms other unsupervised learning-based approaches by up to 28\%. In summary, our contributions are:
\begin{itemize}
    \item A name invariant graph representation for logical formulas that attempts to balance between being faithful to the semantics of these formulas while providing minimal computation overhead.
    \item An ensemble approach for ATPs that leverages different configurations of the underlying un-optimized reasoners to gather more diverse high quality training data.
    \item Experimental evaluation which shows that in comparison to the best unsupervised learning-based reasoners, {\sysname} overall solves more problems, finds most proofs in fewer steps, and has a significantly better transferability especially on hard datasets with the most diverse domains and vocabulary.
\end{itemize}

\sysname's code, data, and trained models are publicly available at \url{https://github.com/ibm/TRAIL}.  
\section{Background}

This work focuses on \emph{saturation-based} theorem proving \cite{robinson1965machine} in first-order logic (FOL) with equality.  In this setting, the automated theorem prover (ATP) is given a \emph{conjecture} (i.e., a formula to be proven true or false) and a logical theory (i.e., an initial set of \emph{axioms} assumed to be true). In addition, the ATP is equipped with a set of \emph{inference rules}, i.e., rules that can be applied to true formulas to yield new true formulas. To execute proof search, the prover proceeds by refutation, i.e., it attempts to show that the negated conjecture together with the axioms entails a contradiction.

In refutation-based theorem proving, the prover adjoins the negated conjecture to the initial set of axioms and applies the inference rules repeatedly to this set until either (1) a contradiction is found, which means that the conjecture is true; or (2) a saturated set is obtained, i.e., a set from which no new formula can be derived (which means the conjecture is false); or (3) its time limit is reached, in which case no conclusion can be derived regarding the validity of the conjecture. A \emph{proof} is a sequence of \emph{inferences} (i.e., inference rule applications) leading from the initial set of formulas to either a contradiction or a saturated set.

The logical theories considered in this work will be theories expressed in the \emph{conjunctive normal form} (CNF). This is the most common representation utilized by state-of-the-art saturation-based theorem provers. A CNF theory consists of a conjunction of clauses, where each clause is a disjunction of literals. The literals within a clause are (possibly negated) atomic formulas, where all variables occurring within the formula are implicitly universally quantified.

\begin{figure*}
\begin{center}
\includegraphics[width=0.8\textwidth]{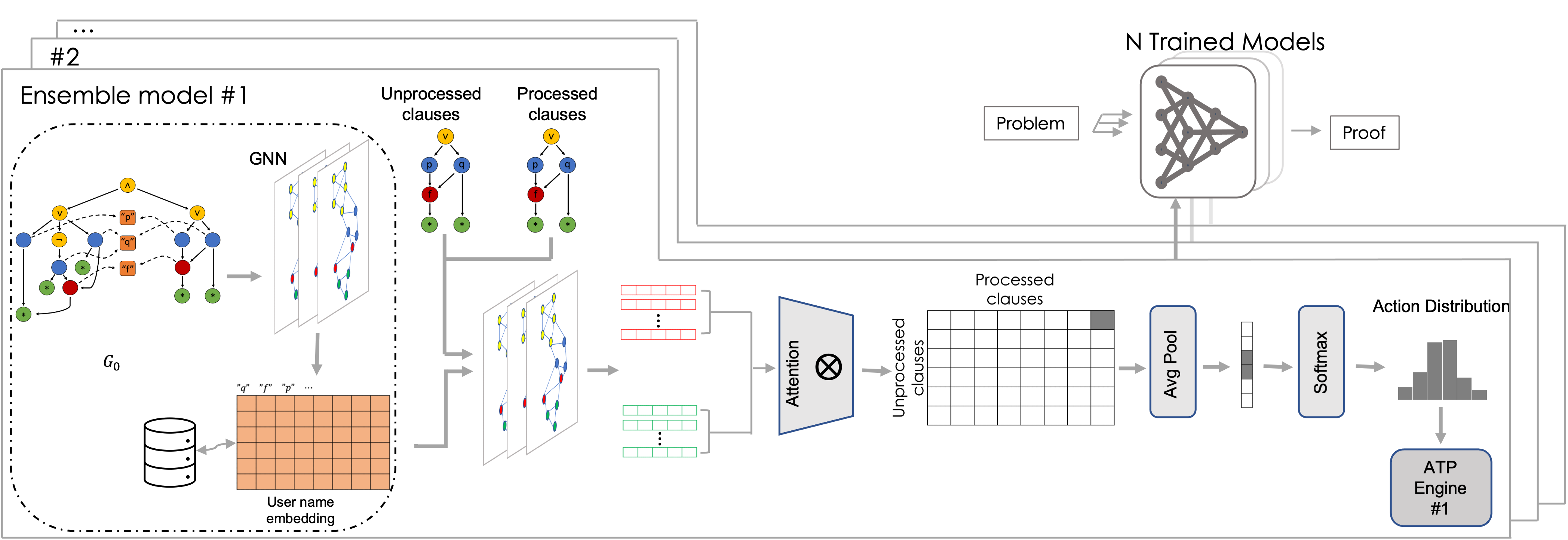}
\caption{{\sysname} Overview: \textit{N} ensembles models are trained, each with their own GNN embedder and policy network. An initial set of node embeddings are computed for the entire input theory, defining a set of contextualized embeddings for user-defined nodes which are later used as the initial embeddings for nodes in individual clause graphs. At inference time, {\sysname} attempts to solve the input problem via the various learned models and output the final proof when found.}
\label{fig:overview}
\end{center}
\end{figure*}

\section{\sysname}

We now describe \sysname, our approach that improves neural proof guidance systems for saturation-based ATPs, e.g., E~\cite{schulz2002brainiac}, Vampire~\cite{kovacs2013first}, etc. Specifically, we present (a) a novel, efficient method for generating name invariant graph neural representations of clauses that provides more meaningful embeddings of non-logical symbols like predicates, functions, and constants and (b) an ensemble method that learns different proof guidance models for different configurations of the underlying ATP. 

As our contributions focus on graph representations and ensemble methods (and not on designing a general neural proof guidance system), we adopt the existing TRAIL~\cite{abdelaziz2022learning} architecture for proof guidance. We emphasize, however, that the method we introduce is not TRAIL-specific. In principle, as it involves replacing the initial embeddings of symbols that serve as the lowest layer of each formula's graph embedding, it is complementary to other high performing unsupervised learning-based proof guidance systems like HER~\cite{aygun2022proving}.

As the details of proof guidance are provided in \cite{abdelaziz2022learning}, we only lightly cover them again here. Figure \ref{fig:overview} shows the entire proof guidance architecture, which is a deep reinforcement learning-based system that employs an attention-based policy network. The state of the theorem prover is represented in terms of the set of processed and unprocessed clauses. The policy network decides which inference to explore based on attention scores computed for pairs of clauses (one from the processed set and one from the unprocessed set). 
During learning, a temperature $\tau$ parameter controls the trade-off between exploration and exploitation, decaying throughout iterations to promote more exploitation over time. Lastly, the reward is computed as the ratio of time taken to solve the problem by the underlying ATP and the time {\sysname} took to solve the same problem. Essentially, this gives the system a higher reward if it found proofs faster than the underlying reasoner. In what follows, we describe {\sysname}'s design and provide the details on our proposed ensembling technique.

\subsection{The Challenge of an Efficient Name Invariant Graph Representation}
Graph-based neural formula representations underlie the most successful deep learning-based proof guidance systems~\cite{abdelaziz2022learning,aygun2022proving}. In these systems, formulas are converted into DAGs (see Figure~\ref{fig:dag_example}) by collapsing identical subtrees of a formula's abstract syntax tree into single nodes with multiple parents. Generally, DAGs are constructed for each individual clause in the provided problem and, during inference, embedded independently of one another with a GNN.

Clause embeddings computed independently can recover some of the dependencies between clauses by using the user-defined names of predicates, functions, and constants to select initial node embeddings; however, such schemes are ultimately limited in that they are leverage name-sensitive representations. Name-sensitive representations can become problematic when testing on a dataset with a completely different vocabulary or, perhaps worse, with a vocabulary that overlaps with the training vocabulary but in which the common names have an entirely different meaning (e.g., a functions $f$ and $g$ in the training dataset have been swapped in the test dataset). 
In principle, predicate, function and constant names are irrelevant artifacts, since the meaning of a given logical theory is unchanged after consistently renaming predicates, functions, and constants. Thus, unnecessarily including specific names in clause representations could result in learning of spurious correlations between those irrelevant artifacts and hinder generalization.

\begin{figure}
\begin{center}
    \begin{subfigure}[b]{0.85\columnwidth}
    \includegraphics[width=0.8\columnwidth]{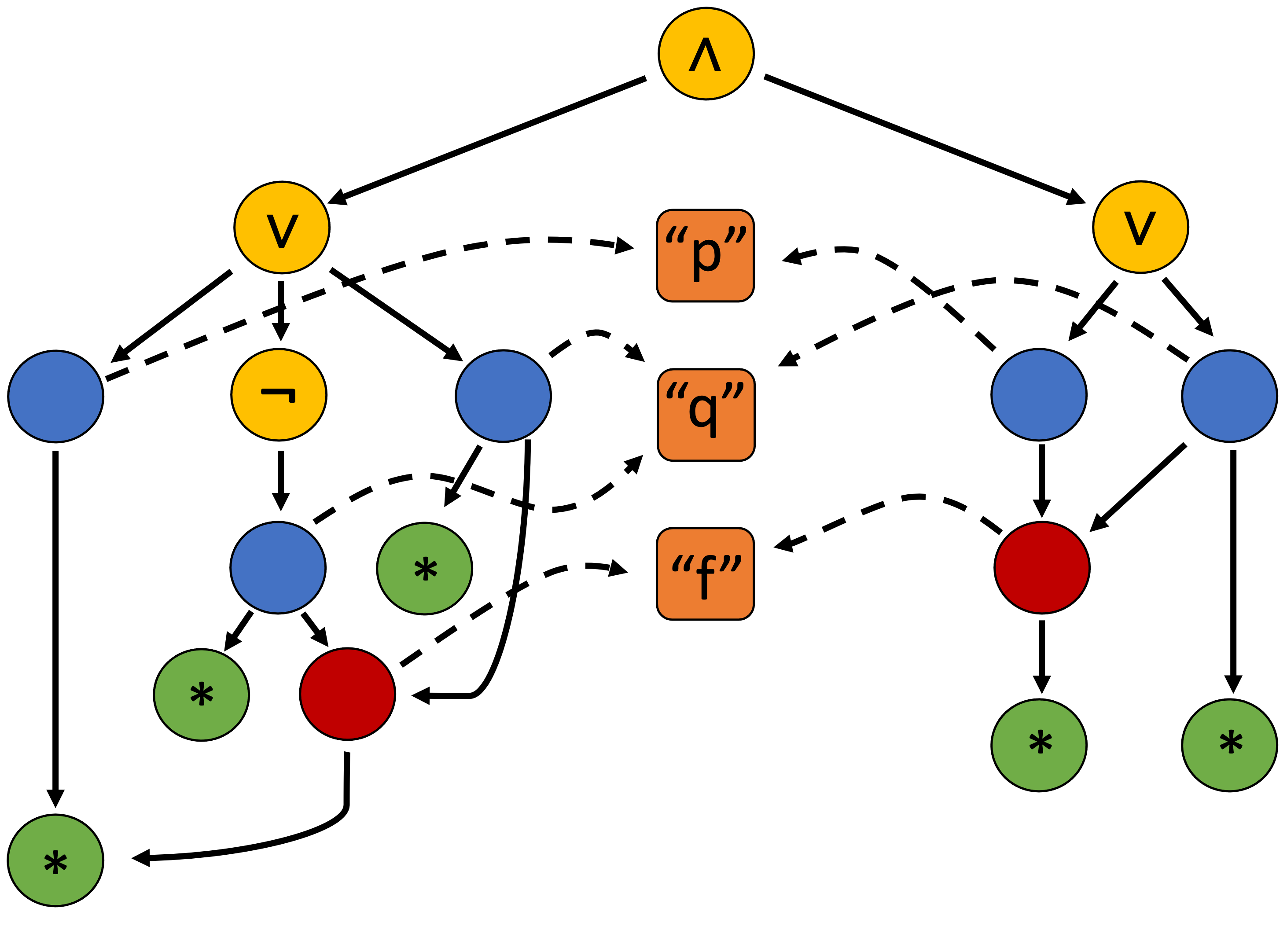}
    \caption{Connected Clauses}
    \label{fig:connected}
  \end{subfigure}
    \vskip 0.5pt
    \begin{subfigure}[b]{0.85\columnwidth}
    \includegraphics[width=0.9\columnwidth]{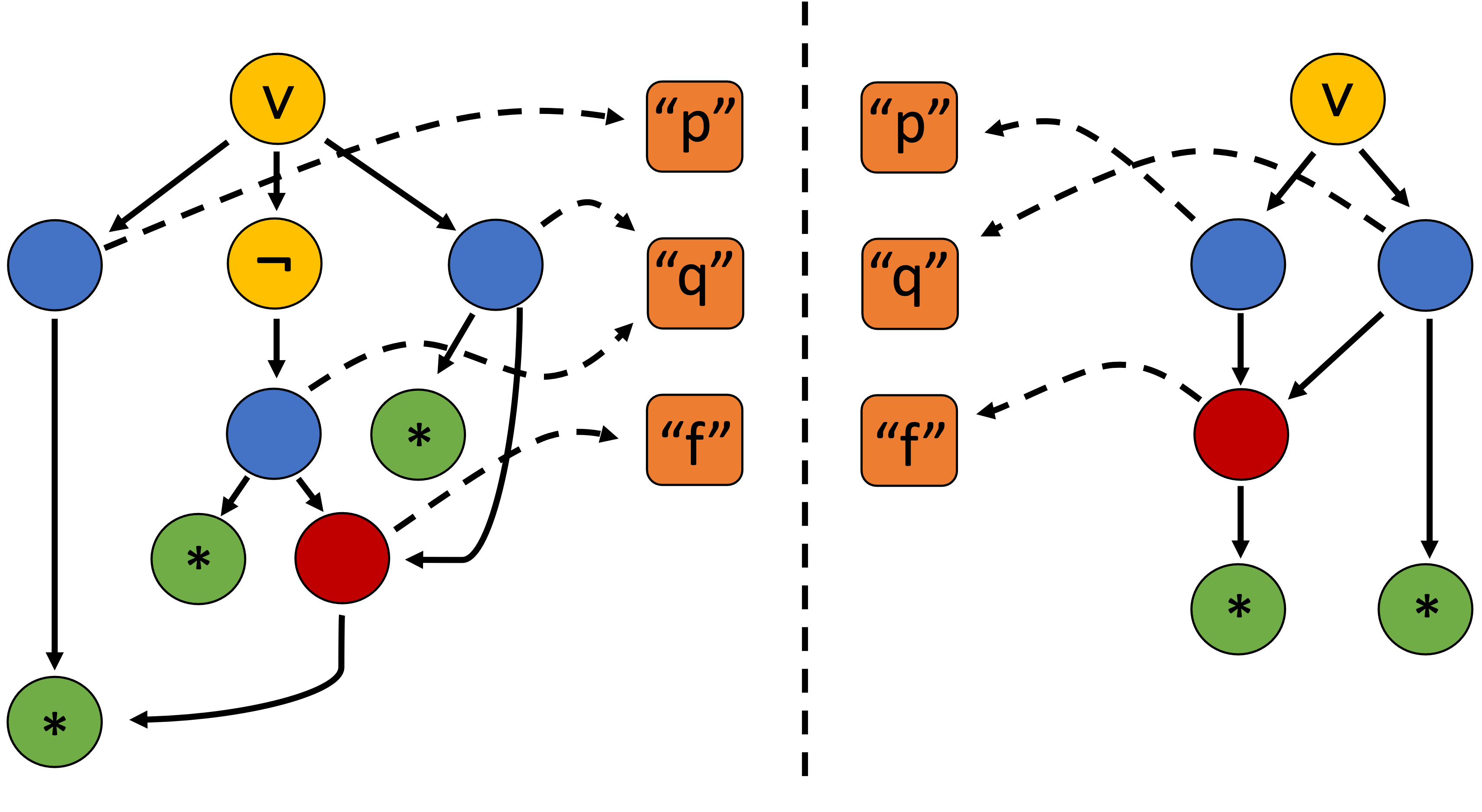}
    \caption{Disconnected Clauses}
    \label{fig:disconnected}
  \end{subfigure}
\caption{Example representation of clauses from Fig~\ref{fig:dag_example} with (a) and without (b) user-defined names sharing. Information about user-defined names  appears only in the square nodes. }
\end{center}
\end{figure}

To avoid name-sensitive representations, we take inspiration from recent works that have explored preserving properties such as invariance to predicate, constant, and function names, and variable quantification order \cite{rawsondirected2020,olvsak2019property,paliwal2020graph}. The system most similar to ours is TRAIL \citet{abdelaziz2022learning,crouse2021deep}, which does not rely on user-defined predicate, function, and constant names. Rather, it produces independent graph embedding using a representation similar to Figure~\ref{fig:disconnected}. In its GNN computation, the initial embedding of a node is only a function of its type (i.e, in the figure this is the color of the node and its shape, but not its label). Unfortunately, this solution has two shortcomings. First, the initial embedding does not preserve the individuality of each symbol (e.g., all square name nodes have the same initial embedding). Second, there is very limited information exchange between disconnected clause graphs. Within a single clause, all applications of the same function or predicate (e.g., connections to the ``q" node in Figure~\ref{fig:disconnected}) have a direct link to the same user-defined name node (e.g., the square node labeled ``q"), however, these square user-defined name nodes are not connected or shared across clause graphs (as shown in Figure~\ref{fig:disconnected}), meaning that the only information they share stems from how they are initialized.

\subsection{Name Invariant Graph Representations}

Our proposed solution is based on the following three key observations:
\begin{enumerate}
    \item  In conjunctive normal form, after an equivalence-preserving variable renaming, the graph connecting all clauses is such that the only shared nodes across clauses are nodes representing the names of constants, functions, and predicates (the square nodes shown in Figure~\ref{fig:connected}). To see why, first note that the universal quantifier distributes over conjunctions. Thus, in the clausal CNF, we can push the top level universal quantifiers inside each clause (i.e., $ [\forall X_1,.., X_n: cl_1 \wedge ... \wedge cl_m] \equiv [(\forall X_1,.., X_n: cl_1) \wedge ... \wedge (\forall X_1,.., X_n: cl_m)] $) and then safely rename variables within each clause in such a way that two distinct clauses do not have any common variables. The only shared symbols between clauses are then constants, functions and predicate names (including skolem constants and functions introduced during the normalization process to remove existentially quantified variables).
    \item All user-defined and skolem names of constants, functions, and predicates are already present in the original normalized theory (with the exception of a select few predicates generated by an infrequently applied E inference strategy).
    \item The meaning of all such constant, function and predicate names is fully specified by the set of initial axioms and the conjecture. 
\end{enumerate}
Based on those three observations, our proposed solution employs the following two-part process:
\begin{enumerate}
    \item  First, before reasoning begins, compute from the fully connected graph of the normalized theory and the negated conjecture (e.g., Figure~\ref{fig:connected}), the embeddings of the vocabulary of predicates, functions, and constants (i.e., square nodes in the figure). This provides meaningful embeddings of non-logical symbols in a way that is name invariant and dependent only on the input theory and the conjecture to prove. 

    \item Second, at each step of reasoning, use the embeddings computed previously as the initial embeddings of named nodes (i.e., square nodes) in the independent graph (e.g., in Figure~\ref{fig:disconnected}) of each clause so that, across all independent graphs, all name nodes with different labels receive different initial embeddings while those with the same label start with the same initial embeddings. In doing so, we address the two key issues of lack of name node individuality and limited information sharing across independent clause graphs.
\end{enumerate}
This procedure has the advantage that it is fairly efficient to compute (requiring only one theory-level embedding computation) while also taking into account much richer structural information than previous embedding methods that isolated each formula. In our implementation, the same GNN model is used both when computing node embeddings for the connected graph of the whole theory and conjecture (see Figure~\ref{fig:connected}) as well as when computing the embeddings for isolated clause graphs such as those in Figure~\ref{fig:disconnected}.

\subsection{Ensemble Method for ATPs}

Ensemble methods are a standard technique used to improve learning performance by combining predictions from multiple models. In the context of learning-based proof guidance systems for ATPs, HER~\cite{aygun2022proving} uses a simple approach that relies on 10 different randomly initialized models to increase the diversity of the search.  However, these different models learn from examples collected from proof attempts using the same configuration of the underlying ATP. Thus, the major source of diversity comes from the random initialization process. 

We first observe that mature ATPs such as E~\cite{schulz2002brainiac} have many configuration options that are orthogonal to the clause selection process that learning-based proof guidance systems such as  HER~\cite{aygun2022proving} or TRAIL~\cite{abdelaziz2022learning} aim to replace. For example, E provides options to select different term order strategies, different literal selection strategies, etc. Those different configurations of the underlying ATP system can provide a greater source of diverse proofs than relying only on the random initialization of multiple models.

Given a time limit $T$ for each proof attempt,  our ensemble approach for ATPs consists of three important steps:
\begin{itemize}
    \item First, we create $N$ different configurations of the underlying ATP to guide (each with their proof guidance or clause selection strategy \textit{disabled}). 
    \item Second, we train, in parallel and independently, each of the $N$ differently configured ATPs with a modified time limit of $T/N$ using, for example, the reinforcement learning based approach of TRAIL  or the Hindsight Experience Replay of HER. In each of the $N$ independent branches, at the start of the first iteration, a randomly initialized model is used. In each branch, at each iteration, attempts to solve all the problems in a dataset are made with the time limit of $T/N$ and the resulting examples are collected to update the model of the branch. The updated model is then used at the next iteration. 
    \item Finally, at the end of each iteration, the set of problems solved is simply the union of problems solved in each of the $N$ independent and parallel branches. 
\end{itemize}

Note that we give equal importance to the $N$ different configurations by dividing the time limit equally among them ($T/N$). This can indeed be extended to give different time limits for different configurations to weight their importance. However, we leave this for our future work.



\section{Experimental Evaluation}

\begin{table}[]
\centering
\begin{tabular}{l|l}
\toprule
Domain           & \#Problems \\
\midrule
Set Theory               & 247           \\
Geometry                 & 207           \\
Kleene Algebra           & 114           \\
Commonsense Reasoning    & 88            \\
Logic Calculi            & 86            \\
Semantic Web             & 79            \\
Relation Algebra         & 48            \\
Processes                & 45            \\
Knowledge Representation & 34            \\
Group Theory             & 25           \\
\bottomrule
\end{tabular}
\caption{Top 10 problems domains in TPTP dataset.}
\label{tab_tptp}
\end{table}

\begin{table*}[]
\footnotesize
\resizebox{2\columnwidth}{!}{%
\begin{tabular}{lll|ll|llllll|l}
\toprule
        &          &            & \multicolumn{2}{c|}{Traditional}                                                                                                                               & \multicolumn{6}{c|}{Unsupervised Learning Based}                                                                                                                                                     &                     \\ 
\midrule
Dataset & Domain   & \#Problems & \multicolumn{1}{c}{\begin{tabular}[c]{@{}c@{}}E \\ (auto-sched)\end{tabular}} & \multicolumn{1}{c|}{\begin{tabular}[c]{@{}c@{}}Vampire \\ (casc)\end{tabular}} & rlCop & plCop & TRAIL & HER   & \begin{tabular}[c]{@{}l@{}}\sysname\\ (10 epochs)\end{tabular} & \begin{tabular}[c]{@{}l@{}}\sysname\\ (1 epoch)\end{tabular} & Stat. Sig. (z-test) \\ 
\midrule
MPTP    & Math     & 2,078      & 63.7                                                                         & \underline{73.0}                                                                          & 31.6   & 42.8   & 58.4 & 68.5 & 72.7                                                                             & \bf 75.5                                                                             & \checkmark ($z = 1.84$)        \\
M2K     & Math     & 2,003      & 97.0                                                                         & \bf 99.0                                                                          & 67.4 & 70.7 & 90.2 & 94.6 & 96.2                                                                            & \underline{98.0}                                                                             & \checkmark ($z = -2.65$)        \\
TPTP    & Multiple & 1,955      & 81.5                                                                         & \bf 84.6                                                                          & -     & -     & -     & -     & -                                                                                & \underline{82.1}                                                                             & \checkmark ($z = -2.15$)        \\ 
\bottomrule
\end{tabular}
}
\caption{Percentage of problems solved across various datasets for multiple traditional and learning-based reasoners. Stat. Sig. indicates whether the improvement over the next best system is statistically significant  ( 'checkmark' for p-value $<$ 0.05).}
\label{tab_conjectures_proven}
\end{table*}

\subsection{Datasets and Baselines}

We evaluate {\sysname} on three benchmarks; \textbf{MPTP}, \textbf{M2K} and \textbf{TPTP}. M2k and MPTP are standard benchmarks that have been used for evaluation by many learning-based systems \cite{aygun2022proving, abdelaziz2022learning, crouse2021deep,zombori2020prolog}. 
Both benchmarks are parts of the larger Mizar \cite{grabowski2010mizar} dataset, with M2K and Mizar consisting of 2,003 and 2,078 Mizar problems, respectively. The main difference between both benchmarks stems from how their constituent problems were selected. M2K problems were selected randomly from the subset of Mizar that is known to be provable by existing ATPs while MPTP problems were selected regardless of whether or not they could be solved by an ATP system. Since problems in both benchmarks are from a single domain (mathematics), we also used TPTP dataset  (Thousands of Problems for Theorem Provers)\footnote{\url{http://tptp.cs.miami.edu/}}. TPTP is the definitive benchmarking library for theorem provers, designed to test ATP performance across a wide range of problem domains (e.g., biology, geography, number theory, etc.). Using TPTP allows us to better test how {\sysname} generalizes to a broader set of domains than would be possible by evaluating with solely Mizar problems.  Table~\ref{tab_tptp} shows the domains comprising the subset of TPTP.

We compare {\sysname} against state-of-the-art traditional and learning-based reasoners. In particular, we compare against the following:
    (1) \textbf{E}~\cite{schulz2002brainiac}:  state-of-the-art theorem prover that relies on manually designed proof guidance heuristics. In our experiments, we used E 2.6 Floral Guranse\footnote{\url{https://wwwlehre.dhbw-stuttgart.de/~sschulz/E/E.html}} which is the latest available version of E (as of August 2022). We ran E prover in \textit{auto-schedule} mode, which is known to solve the most number of problems. 
    (2) \textbf{Vampire}~\cite{kovacs2013first}: another popular theorem prover and the world champion in ATP. We used Vampire version 4.7\footnote{\url{https://github.com/vprover/vampire/}}; the latest version of Vampire when this paper was submitted. We also ran Vampire in best mode, \textit{CASC}, which has competition specific presets for schedule, etc.  
    (3) \textbf{HER} \cite{aygun2022proving}: the most recent learning based approach that uses hindsight experience replay in an incremental learning setting. 
    (4) \textbf{TRAIL} \cite{abdelaziz2022learning} is another recent reinforcement learning based theorem prover that relies on GNNs for logical formula representation. 
    (5) \textbf{rlCop} \cite{KalUMO-NeurIPS18-atp-rl} and (6) \textbf{plCop}~\cite{zombori2020prolog} are two other learning approaches based on reinforcement learning that leverage connection tableau-based theorem proving techniques.

{\sysname} was implemented using E as the underlying reasoner after turning off E's proof guidance/clause selection.
We set the maximum time limit for solving a problem for all systems to \textit{100 seconds}. We ran E and Vampire ourselves on the same hardware setup used for {\sysname} and used the numbers reported in TRAIL and HER papers (both systems used a 100-second time limit as well). Note that the numbers we obtained for E prover are comparable to those reported by HER and better than those reported by TRAIL. Detailed information about software \& hardware used is provided in supplemental materials.

\subsection{Number of Problems Solved}
In this experiment, we compare the number of problems solved by {\sysname}  to other baselines within the allowed time limit. Table \ref{tab_conjectures_proven} shows the results for each system (including two variants of {\sysname} where the model update step after each iteration is done in either 1 or 10 epochs, respectively). The state-of-the-art traditional reasoners, E and Vampire, were able to solve 63.7\%(1,324) and 73.0\% (1,517) problems on the harder MPTP dataset, respectively. 
In comparison, learning-based approaches managed to provide competitive performance, with HER and {\sysname} solving 5\% and 12\% more problems compared to E (a relative improvement of 7\% and 18\%), respectively.
The use of our ensemble method and the name-invariant representation of clauses allowed {\sysname} to outperform HER by 10\% (1,569 vs. 1,424).
The same behavior can be observed on the M2K dataset with {\sysname} being the only  learning-based system that outperforms E, and only 20 problems away from Vampire. Similarly, on TPTP, {\sysname} slightly outperformed E by proving 12 more conjectures. This is somewhat unsurprising, as TPTP is a particularly hard dataset due to its varied domains that each contain smaller numbers of problems (thus making it a valuable benchmark to test on). Although {\sysname} slightly underperforms compared to Vampire in TPTP and M2k, it provides a very competitive performance which allowed it to outperform E across all datasets and Vampire on the hard Mizar dataset. We believe this is a very promising performance for a system that does not rely on any heuristics and is  learning completely from scratch. 

\begin{figure}[tb!]
\begin{center}
\includegraphics[width=0.9\columnwidth]{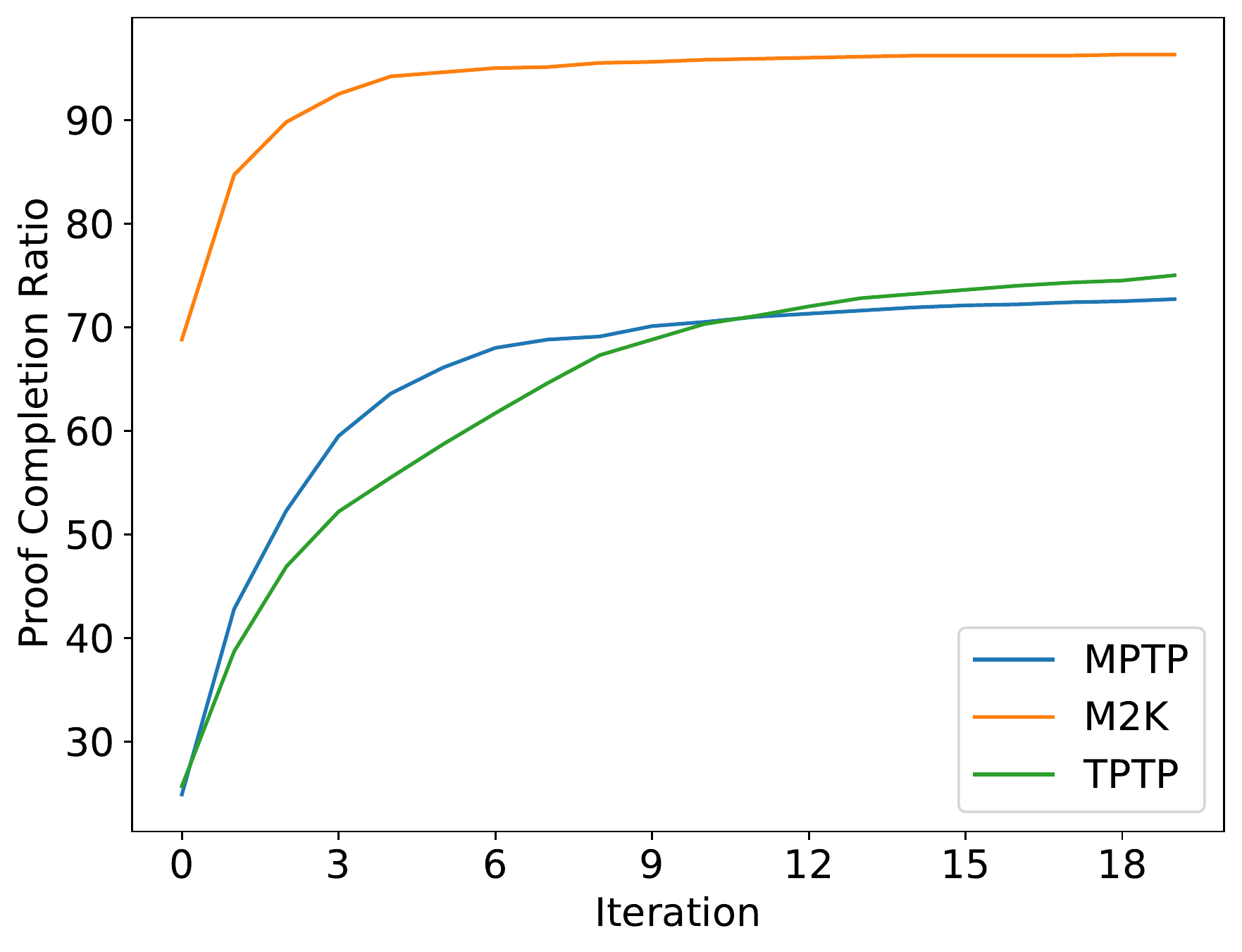}

\caption{Percentage of problems solved across iterations}
\label{fig:compl_ratio}
\end{center}
\end{figure}

We also show in Figure \ref{fig:compl_ratio} the progress of {\sysname} in terms of the percentage of problems solved across iterations. In all three datasets, the majority of the improvement happens in the first ten iterations, afterwards the model keeps improving but slowly. Compared to MPTP and TPTP, M2K model initial performance (iteration 0) is much higher (68.8) due to the relative simplicity of the dataset. 

Finally, although we use the same 100 secs time limit as other tabula rasa approaches (HER, TRAIL \& mlCOP), with our ensemble approach, we divide the overall time budget among 4 models, i.e., 25 secs per model. Only a small fraction of that time limit is used to solve the vast majority of problems. For example,  on MPTP, all 4 models solved individually, in 25 secs, more problems than E with 100 secs: 1,343, 1,388, 1,409, and 1,361 compared to 1,324 by E. Furthermore, by the end of 20 iterations (for models trained with 10 epochs), most problems are solved in few secs. On MPTP, one of the models used an average of 1 sec per problem   solved (stddev: 1.6). Thus, the overhead introduced by our approach remains small.

\begin{table*}[]
 \resizebox{\textwidth}{!}{%
\begin{tabular}{l|c|c|c|c|c|c}
\toprule
& TPTP $\rightarrow$  MPTP & TPTP $\rightarrow$ M2k & MPTP $\rightarrow$ M2k & MPTP $\rightarrow$ TPTP & M2k $\rightarrow$ MPTP & M2k $\rightarrow$ TPTP \\
\midrule
{\sysname}  & \textbf{48.7}&  \textbf{86.6} &    \textbf{93.5}   &   \textbf{50.8}  &     \textbf{64.5} &  \textbf{53.6}                         \\
{\sysname} (Ensemble Only)
&         39.1               &           82.0 &                  90.4      &      47.2                   &     59.1  &    50.6               \\
TRAIL  & -    & - &    83.5   &  - &     50.4           &  -    \\
\bottomrule
 Stat. Sig.(z-test) &  \checkmark ($z = 6.25$)&  \checkmark ($z = 4 $) & \checkmark ($z = 3.61$) & \checkmark ($z = 2.25$) & \checkmark ($z = 3.58$) &  \checkmark ($z =1.88 $) \\
\bottomrule
\end{tabular}
}
\caption{Transfer learning: training on one dataset and testing on another. The numbers are the \textit{percentages} of solved problems. "Ensemble Only" refers to {\sysname} operating \emph{without} the theory-level graph embeddings.
}
\label{tab_transfer}
\end{table*}

\subsection{Transfer Learning}
Table \ref{tab_transfer} shows {\sysname}'s performance when trained on one dataset and tested on another. Note that, while MPTP and M2K are both from the same mathematics domain, the TPTP problems are drawn from a different, varied set of domains (see Table \ref{tab_tptp}). As expected, the number of problems solved is maximized when training and testing on the same domain. 
However, it can be seen that transfer learning is happening, e.g., the M2k-trained model was able to solve 53.6\% TPTP problems in testing mode. Similarly, the TPTP-trained model showed promising performance when tested on MPTP and M2k datasets where it solved 48.7\% (1,013) and 86.6\%(1,736) of the problems, respectively. Importantly, Table~\ref{tab_transfer} includes results for NIAGRA without the name invariant graph embeddings (i.e., only the ensemble scheme remained) in the second row. The results clearly show that {\sysname} with our efficient name invariant graph embeddings (first row) is superior to {\sysname} without it (second row). In addition, as compared to TRAIL (the only prior system reporting transfer learning results between MPTP and M2K), our method achieves significantly improved performance.


\subsection{Ablation Experiments}

\begin{table}[]
\begin{small}
\centering
 \resizebox{\columnwidth}{!}{%
\begin{tabular}{lcc|l}
\toprule
Dataset     & \multicolumn{1}{c}{\begin{tabular}[c]{@{}c@{}}{\sysname} \\ (Ensemble Only)\end{tabular}} & {\sysname} & Stat. Sig. \footnotesize{(z-test)} \\
\midrule
MPTP &        67.6           &     \textbf{72.7}  & \checkmark ($z = 3.59$)          \\
M2K  &        94.6            &     \textbf{96.2} & \checkmark ($z = 2.43$)             \\
TPTP &      66.3             &     \textbf{78.9}  &   \checkmark ($z = 7.79$)       \\
\bottomrule
\end{tabular}
}
\end{small}
\caption{Effect of faithful graph representation: {\sysname}'s performance (percentage of problems solved) with and without the initial whole theory graph.}
\label{init_graph_embed}
\end{table}

\subsubsection{Effect of Faithful Graph Representation}
In this experiment, we show the effect of switching on and off the initial graph embedding. Our hypothesis is that having the full theory graph where user-defined names are shared allows {\sysname} to learn meaningful embeddings of non-logical symbols in a way that is both name invariant and dependent on the input theory. Table \ref{init_graph_embed} shows the performance of {\sysname} with and without (ensemble only) the connected graph of the whole theory.  It can be seen that the initial embeddings obtained from the normalized theory graph helped {\sysname} to achieve significantly better performance (relative improvement) on the two hardest datasets: MPTP (+7.5\%)  and TPTP (+17\%). As expected, our efficient name invariant scheme has the most significant impact (17\% improvement) on the TPTP dataset, which uses the most diverse vocabulary and domains.

\subsubsection{Ensemble Size} We also evaluated our ensemble  against both non-ensemble methods and methods employing a standard ensemble setting (e.g. in HER) where multiple learners are used which only differ in their  initial random model weights. Our results show that our ensemble approach performs much better overall. Due to lack of space, we add the details of this experiment to the supplementary material.

\section{Related Work}



Reinforcement learning (RL) has been used to learn for proof-guidance heuristics from scratch. 
For example, \cite{KalUMO-NeurIPS18-atp-rl,zombori2020prolog} combine RL with Monte Carlo tree search to guide FOL tableau-based ATPs.
\citeauthor{bansal2019icml} (\citeyear{bansal2019icml}) refer to deep RL-based interactive theorem proving in HOL Light and use imitation learning with an application to higher-order logic. \citeauthor{aygun2022proving} (\citeyear{aygun2022proving}) apply the idea of hindsight experience replay (HER) of \citeauthor{andrychowicz2017neurips} (\citeyear{andrychowicz2017neurips}) to ATP. 
TRAIL \cite{abdelaziz2022learning} leverages deep RL to learn how to guide a saturation-based theorem prover. It uses a graph neural network to represent a proof state as well as an attention-based policy network to efficiently learn interactions between clauses and actions. 
Unlike these methods, {\sysname} is based on two key contributions 1) capturing the name-invariant theory-level structural and semantic information when embedding logical clauses and 2) an ensemble approach that uses different configurations of the underlying unoptimized reasoner. 

Other relevant non-RL-based approaches using deep learning for proof guidance include \cite{loos2017lpar,paliwal2020graph}. 
These methods use neural networks to select inferences, but, as mentioned before, use local formula representations that exclude theory-level structural information. In addition, the work of \cite{suda2021vampire} used an RNN to process the derivation history of a clause to determine its relevance for proof search. The ENIGMA system of \cite{jakubuuv2020enigma} uses the symbol-independent GNN introduced in \cite{olvsak2019property} to embed clauses for relevance scoring. As mentioned in the introduction, their approach embeds batches of clauses taken from the unprocessed set together and computes a single fixed relevance score for each clause. Thus, while their method may result in clause representations that are more reflective of the current overall state of a theorem prover, it has the disadvantage of requiring larger-scale operations to be performed more frequently (in contrast to our work, which executes a theory-level embedding operation only once).


Our work on ensembling is related to so-called algorithm portfolios \cite{huberman1997science}. When solving combinatorial search problems, algorithm portfolios consist of different search algorithms or identical algorithms with different parameter configurations. 
Having a variety of portfolios allows to examine different portions of the search spaces. While the performance of the algorithm portfolios has been tested in various domains such as SAT solving, puzzle solving and domain-independent planning \cite{hamadi2009sat,valenzano:icaps2010,katz2018ipc}, its empirical efficiency has not been explored in unsupervised learning-based ATPs.

\section{Conclusion}
In this paper, we proposed {\sysname}, a novel graph embedding approach for automated theorem proving aimed at finding a better balance in the trade-off between semantic-faithfulness and computational overhead. In addition, we contributed an ensemble technique for theorem proving that, unlike prior works using ensembles of identical models with different weight initializations, trains models with different configurations of the underlying unoptimized reasoner. 
Our results on datasets from various domains shows that NIAGRA achieves state-of-the-art performance outperforming the best learning-based reasoners by solving up to 10\% more problems. It also provides  comparable and sometimes better performance than the best heuristics-based reasoners, despite learning proof strategies completely from scratch.  
It also exhibits better transfer on hard datasets with diverse domains and vocabulary compared to other learning-based proof guidance systems. 

\bibliographystyle{named}
\bibliography{aaai23}

\end{document}


\maketitle

\section{Supplementary Material}


\subsection{Hardware Setup}
Our experimental setup uses CPU and GPU machines running Red Hat Enterprise Linux release 8.5 (Ootpa). CPU machines are used during the proof search process; i.e. compute the embeddings, run it through the policy network, and perform the reasoning step via the underlying reasoner. GPU machines are used for model training, after each iteration, collected examples are shipped to the GPU server which loads the latest checkpoint and use the examples to update its parameters. Our CPU machines have 16 CPU cores and a total of 240GB of RAM. GPU machines on the other hand runs an NVIDIA A100 Tensor Core GPU with 40GB of GPU memory. We also ran E and Vampire on the same setup, but on the CPU machines since they don't make use of the GPU. 

\begin{table}[tbh!]
\centering
\begin{tabular}{ll}
\toprule
Hyperparameter & Value   \\ 
\midrule
    $\tau$ (temperature)              &   3       \\  
    Dropout                     &   0.57     \\
 Learning rate               &   0.001   \\
    Temp. decay                  &   0.89.   \\  
    $\lambda$ (regularization.)            &   0.004 \\    
    Epochs                  &  10     \\
    Min reward              &  1.0       \\   
    Max reward              &   2.0     \\  
    Activation function &  ReLU    \\
    Reward normalization        &   normalized    \\ 
    GCN conv. layers        &  2     \\
\bottomrule
\end{tabular}
\caption{Hyperparameter values }
\label{tab:hyperparams_tunable}    
\end{table}

\subsection{Hyper-parameter values}

Table~\ref{tab:hyperparams_tunable} shows the values of hyper-parameters used in our experiments.

\subsection{Number of reasoning steps}
\begin{figure*}[tbh!]
\centering
  \begin{subfigure}[b]{0.33\textwidth}
    \includegraphics[width=\textwidth]{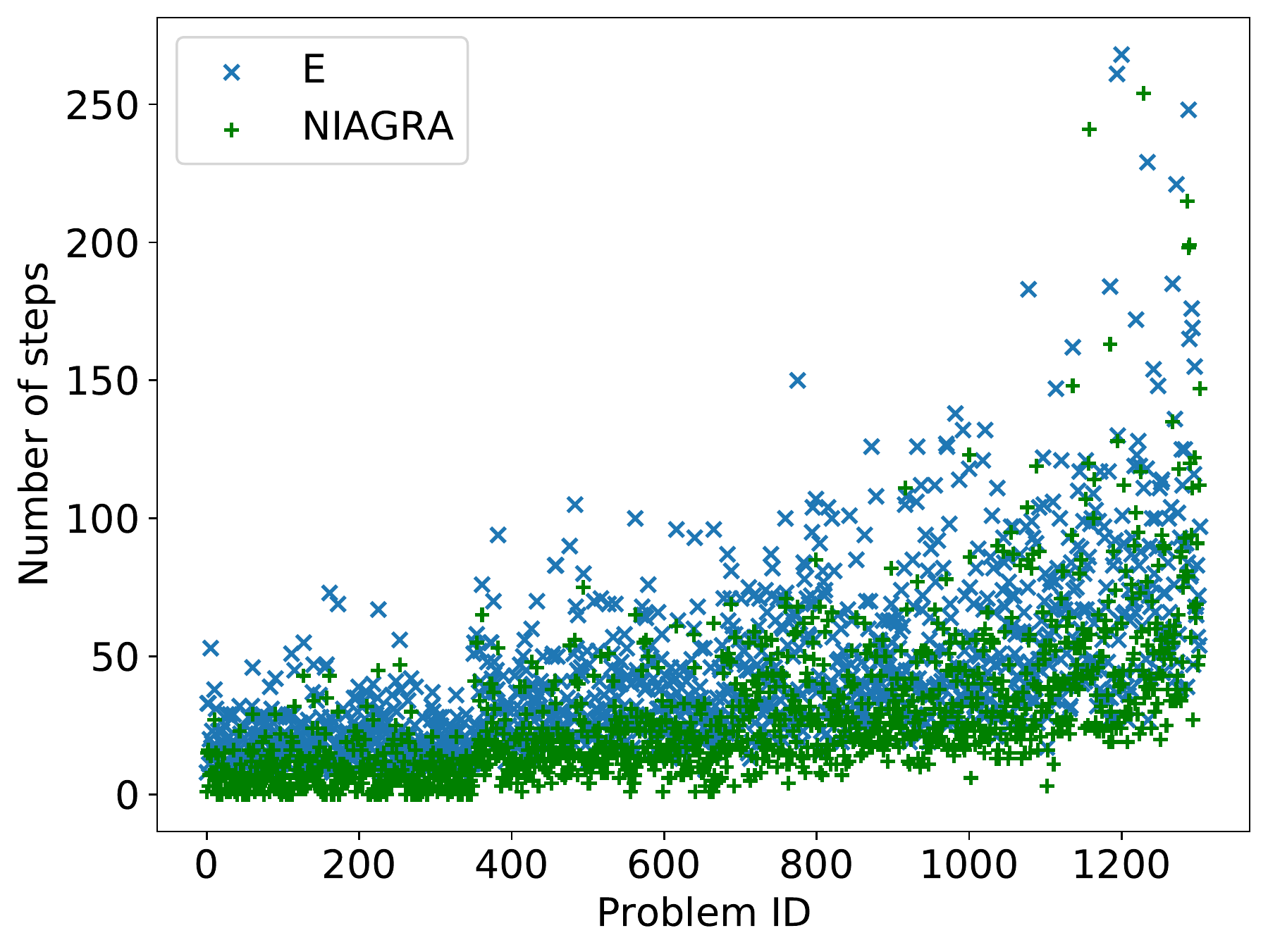}

    \caption{MPTP}
    \label{fig:1}
  \end{subfigure}\hfill
  \begin{subfigure}[b]{0.33\textwidth}
    \includegraphics[width=\textwidth]{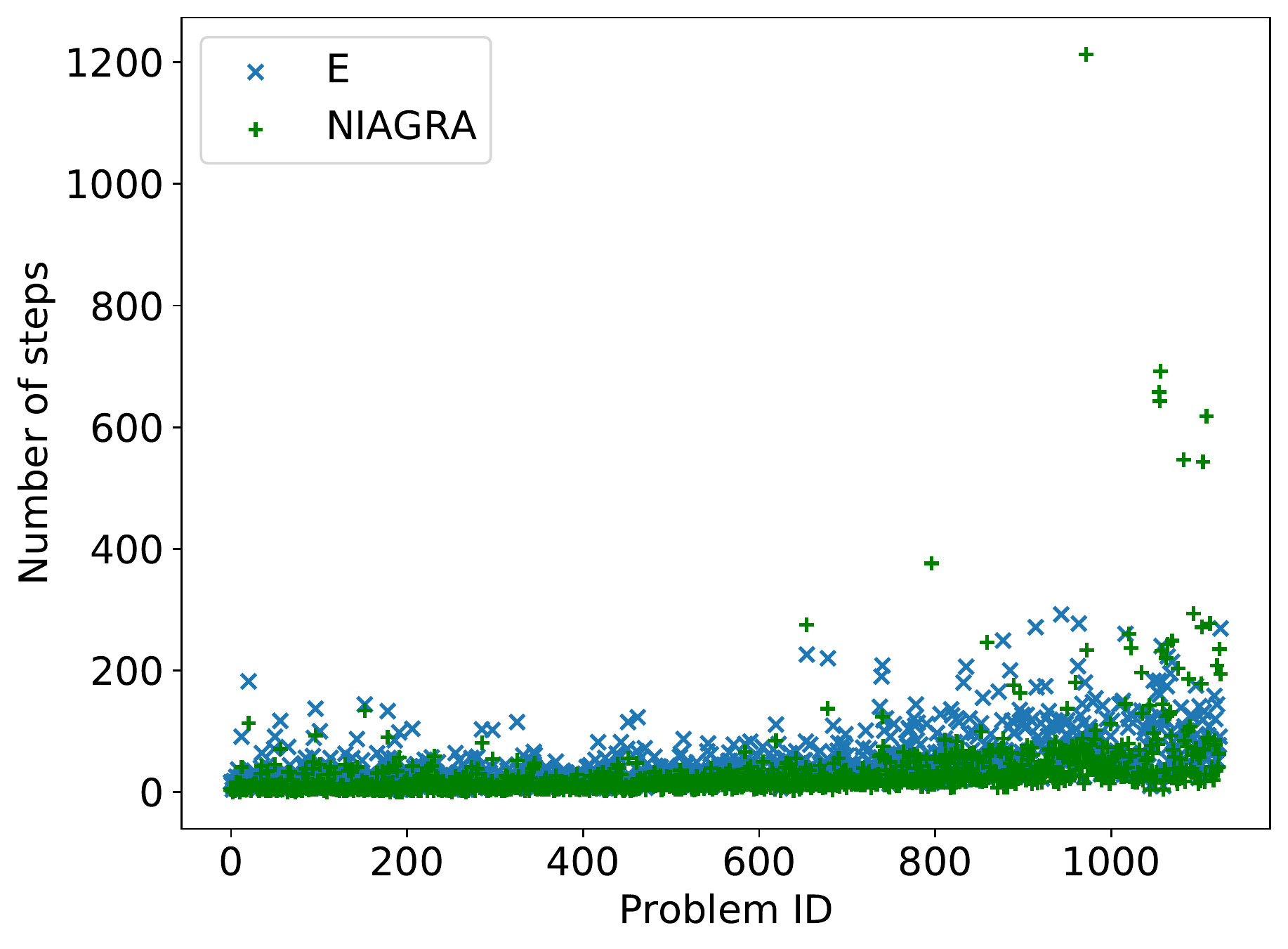}

    \caption{TPTP}
    \label{fig:2}
  \end{subfigure}
  \begin{subfigure}[b]{0.33\textwidth}
    \includegraphics[width=\textwidth]{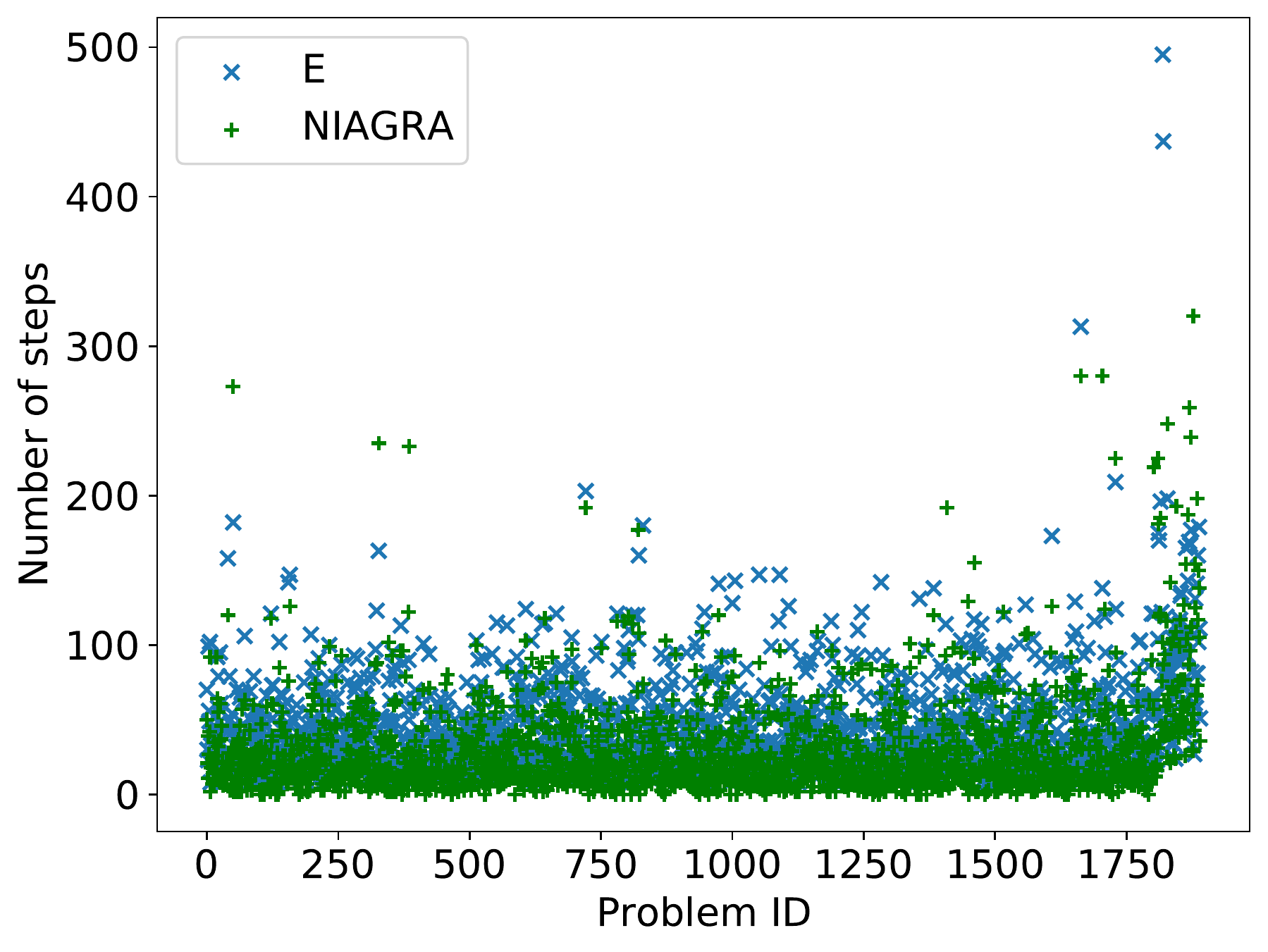}

    \caption{M2K}
    \label{fig:2}
  \end{subfigure}
   \caption{Number of reasoning steps for {\sysname} in comparison with E for MPTP and TPTP datasets}
    \label{fig:proof_len}
\end{figure*}

Figure \ref{fig:proof_len} shows the number of reasoning steps taken for the subset of problems solved by both {\sysname} and E. On MPTP dataset, {\sysname} takes fewer number of steps than E in 1,261 problems (96.7\%) whereas E manages to conclude on a conjecture in fewer steps than {\sysname} in only 40 problems. Similar results were obtained on the TPTP dataset, where {\sysname} reaches a conclusion in fewer steps for 1,064 problems (94.6\%) versus E for 60 problems only. On M2k, {\sysname} concluded in fewer steps in 1,796 problems (94.9\%) when E did better in 89 problems only.
On M2K, {\sysname} managed to conclude a conjecture in fewer steps than E in 1,766 problems (94.9\%) whereas E solved 89 problems (4.9\%) in fewer steps than {\sysname}. 

\begin{table*}[h!]
\centering
\begin{tabular}{l|llll}
\toprule
     Dataset
     & \multicolumn{1}{c}{\begin{tabular}[c]{@{}c@{}}1\\ (100s each)\end{tabular}} & \multicolumn{1}{c}{\begin{tabular}[c]{@{}c@{}}2 \\ (50s  each)\end{tabular}} & \multicolumn{1}{c}{\begin{tabular}[c]{@{}c@{}}4 \\ (25s each)\end{tabular}} &
     \multicolumn{1}{c}{\begin{tabular}[c]{@{}c@{}}4 \\ (25s each)\\ (same E config) \end{tabular}} \\
\midrule
E's configuration & Auto0 & Auto0, Auto1 & \multicolumn{1}{c}{\begin{tabular}[c]{@{}c@{}}Auto0, Auto1, Auto2,\\ SelectLargestNegLit \end{tabular}}  & SelectLargestNegLit \\
\midrule
MPTP &    67.1    &  69.1 &  \textbf{70.1}   & 68.1 \\
M2K  & 87.2$^*$     &   95.0     &  \textbf{95.6}    &  93.8  \\
TPTP &  65.5  &    71.8     & \textbf{73.2}    & 65.4  \\
\bottomrule
\end{tabular}
 \caption{Effect of ensemble sizes on performance  (as \textit{percentage}): in all setups, the overall time limit is 100 seconds. $^*$ M2K ensemble size of 1 crashed due to memory limit and finished only 3 learning iterations. Due to resource constraints, we could not run all variations to the same number of iterations and hence capped the performance of all variations to 10, 10 and 15 iterations for MPTP, M2K and TPTP, respectively.}
 \label{tab:ensemble}
\end{table*}

\subsection{Ensemble Method}
In this experiment, we show the performance of {\sysname} as we change the size of the ensemble. In particular, we vary the number of models we train between 1, 2 and 4 models. In all setting, the total time limit allowed for all models collectively is 100 seconds where we divide the time given to each model equally. Table \ref{tab:ensemble} shows the results. The best performance is obtained with 4 ensemble models, each allowed a 25 seconds time limit. We also compare to the standard setting of using an ensemble of size 4, but using the same config of the underlying unoptimized reasoner. This is a similar setting to HER  where they used multiple learners, where the only difference is in the initial random model. As Table \ref{tab:ensemble} shows, our ensemble approach shows much better performance to this standard approach across all datasets. Table~\ref{tab:ensemble} also reveals that running 4 models (trained on 4 different configurations of E) for 25 seconds each is superior to running a single model for 100 seconds.
